\documentclass[11pt]{article}
\usepackage[utf8]{inputenc}
\usepackage{amsmath}
\usepackage{amsfonts}
\usepackage{amssymb}
\usepackage{graphicx}
\usepackage{booktabs}
\usepackage{url}
\usepackage[pdftex,colorlinks=true,linkcolor=blue,citecolor=blue,urlcolor=blue]{hyperref}
\hypersetup{pdfinfo={
    Title={ChexFract: From General to Specialized - Enhancing Fracture Description Generation},
    Author={Nikolay Nechaev, Evgeniia Przhezdzetskaia, Dmitry Umerenkov, Dmitry V. Dylov},
    Subject={Medical AI, Radiology, Fracture Detection},
    Keywords={medical AI, radiology, fracture detection, vision-language models, AIRI}
}}
\usepackage{natbib}
\usepackage{geometry}
\title{ChexFract: From General to Specialized - Enhancing Fracture Description Generation}

\author{
    Nikolay Nechaev$^1$ \\
    \texttt{nechaev@airi.net}
    \and
    Evgeniia Przhezdzetskaia$^1$ \\
    \texttt{przhezdzetskaia@airi.net}
    \and
    Dmitry Umerenkov$^1$ \\
    \texttt{dumerenkov@airi.net}
    \and
    Dmitry V. Dylov$^{1,2}$ \\
    \texttt{d.dylov@gmail.com}
    \\[8pt]
    $^1$Artificial Intelligence Research Institute (AIRI), Moscow, Russia \\
    $^2$Skolkovo Institute of Science and Technology (Skoltech), Moscow, Russia
}

\date{} 

\begin{document}

\maketitle

\begin{abstract}
Generating accurate and clinically meaningful radiology reports from chest X-ray images remains a significant challenge in medical AI. While recent vision-language models achieve strong results in general radiology report generation, they often fail to adequately describe rare but clinically important pathologies like fractures. This work addresses this gap by developing specialized models for fracture pathology detection and description. 
We train fracture-specific vision-language models with encoders from MAIRA-2 and CheXagent, demonstrating significant improvements over general-purpose models in generating accurate fracture descriptions. Analysis of model outputs by fracture type, location, and age reveals distinct strengths and limitations of current vision-language model architectures. We publicly release 
our best-performing fracture-reporting model, facilitating future research in accurate reporting of rare pathologies.
\end{abstract}


\section{Introduction}

Radiology reports are critical for clinical decision-making, yet automated generation of accurate reports from chest X-rays (CXRs) remains challenging, particularly for rare but clinically significant pathologies such as fractures. Missed or inaccurately described fractures can lead to delayed diagnosis, inappropriate treatment, and poorer patient outcomes, underscoring the need for specialized automated solutions. Existing vision-language models (VLMs) like MAIRA-2 and CheXagent achieve impressive results in general radiology report generation but struggle to reliably detect and describe fractures due to their scarcity in available datasets and inherent complexity in radiology descriptions.

The primary objective of this work is to address these shortcomings by creating a specialized fracture-specific dataset and developing optimized fracture-reporting vision-language models to enhance clinical reporting accuracy. We demonstrate how fracture-focused fine-tuning and careful encoder selection substantially enhance model performance in identifying and describing fractures. By publicly releasing 
our optimized fracture-reporting models, we aim to enable more accurate clinical reporting of rare yet important pathologies.

\section{Related Work}

The generation of radiology reports from chest X-ray (CXR) images has garnered significant attention in recent years, with advancements in both classification and text generation models.

\subsection{Vision-Language Models for Radiology Report Generation}

Recent developments in vision-language models (VLMs) \cite{li2025benchmark} have aimed to generate comprehensive radiology reports. MAIRA-2 \cite{maira2} is a large multimodal model designed for grounded radiology report generation, combining a radiology-specific image encoder with a large language model (LLM) to generate chest X-ray reports with and without grounding. Similarly, CheXagent \cite{chen2024visionlanguagefoundationmodelenhance} is an instruction-tuned foundation model capable of analyzing and summarizing CXRs, integrating a clinical LLM for parsing radiology reports and a vision encoder for representing CXR images.

While these models achieve strong results on general report generation benchmarks, they often underperform in accurately describing rare but clinically significant pathologies, such as fractures. This limitation underscores the need for specialized approaches targeting specific pathologies to enhance the clinical utility of automated report generation systems.

\subsection{Classification Models in Chest X-ray Analysis}

Before vision-language models, convolutional networks like CheXNet \cite{rajpurkar2017chexnet}, a 121-layer convolutional neural network based on DenseNet-121 \cite{huang2017densenet} and COVID-Net \cite{wang2020covid} demonstrated strong performance in thoracic disease classification, achieving radiologist-level accuracy in pneumonia and COVID-19 detection.

Despite their success in classification tasks, these models are unable to generate descriptive reports, which are essential for comprehensive clinical assessments. This gap has motivated the integration of classification strengths into VLMs to enhance report generation capabilities.

\subsection{Datasets for Chest X-ray Report Generation}

The development and evaluation of both classification and report generation models heavily rely on large-scale, annotated datasets. The MIMIC-CXR dataset \cite{johnson2019mimic} is a publicly available resource comprising over 370,000 chest radiographs with corresponding free-text radiology reports, facilitating research in automated report generation. PadChest \cite{bustos2020padchest} is another extensive dataset containing more than 160,000 high-resolution CXR images with multi-label annotations and associated reports, supporting multi-label classification and report generation tasks. Additionally, the Open-I dataset \cite{demner2011preparing} from Indiana University provides a collection of chest X-ray images paired with radiology reports, serving as a valuable resource for developing and benchmarking report generation models.

Despite their utility in advancing automated radiology report generation, existing datasets exhibit significant class imbalance, with a predominance of non-critical or frequently occurring findings and a relative underrepresentation of clinically important conditions such as fractures. This imbalance is further exacerbated by the intrinsic scarcity of annotated fracture cases, which limits the model's ability to effectively learn and generalize to these rare pathologies. Consequently, the development of targeted datasets with enriched representation of underdiagnosed yet clinically relevant abnormalities is essential to improve model robustness and diagnostic accuracy. ChexFract specifically addresses this gap by providing a large-scale, fracture-focused dataset, enriching representation of underdiagnosed yet clinically critical pathologies to significantly enhance model generalization and diagnostic accuracy.

\subsection{MIMIC-CXR Test Set Relabeling}

\subsubsection{Motivation and Rationale}

During our preliminary analysis, we identified that the original fracture labels in the MIMIC-CXR test set -- generated using the CheXpert \cite{irvin2019chexpert} labeler, which serves as the standard classification ground truth (GT) for this dataset -- were often suboptimal. CheXpert, a rule-based system that relies on keyword matching, frequently fails to accurately capture fractures described using nuanced language, synonyms, or complex contextual phrasing in free-text radiology reports. As a result, it introduces a substantial number of false negatives, missing fractures that are clearly mentioned by radiologists.

To create a more reliable and semantically accurate test set, we decided to perform a complete relabeling of MIMIC-CXR test set using a GPT-4o LLM. Our rationale was that GPT-4o could interpret reports far more accurately than rule-based systems. We aimed not only for a binary "fracture/no fracture" classification but for an enriched annotation that included:

\begin{enumerate}
    \item \textbf{A three-class label:}
    \begin{itemize}
        \item \textit{Fracture:} An explicit mention or description of a fracture.
        \item \textit{Normal:} An explicit statement confirming the absence of traumatic changes.
        \item \textit{Other:} No information regarding fractures. This distinction is crucial for separating reports that ruled out fractures from those that did not mention them.
    \end{itemize}
    \item \textbf{Detailed attributes:} Extraction of granular information about the fracture, including:
    \begin{itemize}
        \item \textit{Location:} Ribs, Clavicle, Shoulder, Spine, Sternum, Scapula, Sternal Wires or Other.
        \item \textit{Side:} Left, Right, Both or None.
        \item \textit{Stage:} Acute, Healed or Other.
        \item \textit{Implants:} Presence of Screws, Rods, Plates or Other.
    \end{itemize}
\end{enumerate}

\subsubsection{Relabeling Process}

The relabeling process was automated using a script that leveraged the OpenAI Batch API. For each report in the MIMIC-CXR test set, we followed a systematic pipeline to enhance label quality. First, we employed prompt engineering to design a system prompt that instructed GPT-4o to act as an experienced radiologist. This prompt included detailed guidelines for classifying the report, extracting fracture-related attributes, and citing the specific text snippet that supported the classification.

Next, the report text was submitted to the GPT-4o model through an API request, with the requirement to return a structured JSON output. The use of a predefined JSON schema ensured that the output was both consistent and valid across all reports.

Finally, the structured responses generated by GPT-4o were aggregated into a single CSV file. This formed our new, relabeled "gold standard" dataset for evaluating the performance of report generation models in describing fractures.

This methodology enabled us to construct a more accurate and fine-grained test set. Additionally we manually validated all the cases where the CheXpert and GPT-4o labels disagreed. While the original CheXpert labeling identified 77 fracture cases within the 2,921 reports of the MIMIC-CXR test set, our GPT-4o-based relabeling identified 154 cases. This demonstrates a substantial improvement in sensitivity. Examples of reports where CheXpert failed to detect fractures that were correctly annotated by GPT-4o are presented in Table \ref{tab:relabeling_examples_combined}.

\begin{table*}[t]
\centering
\begin{tabular}{p{1.7cm} p{3.0cm} p{2.8cm} p{7.3cm}}
\toprule
\textbf{Study ID} & \textbf{CheXpert label (Fracture)} & \textbf{GPT-4o label (Fracture)} & \textbf{GPT-4o Quote} \\
\midrule
59981256 & NaN & 1.0 & There are chronic rib fractures. \\
56618763 & NaN & 1.0 & Bilateral rib fractures are noted. \\
59968351 & NaN & 1.0 & Stable mid-thoracic compression fracture. \\
51830719 & NaN & 1.0 & An old left clavicular deformity is noted. \\
54759244 & 1.0 & 0.0 & No displaced fracture is seen. \\
59041431 & 1.0 & 0.0 & No fracture is visualized. \\
53452091 & 1.0 & 0.0 & There are no displaced rib fractures. \\
59454336 & 1.0 & 0.0 & No displaced fracture is seen. \\
\bottomrule
\end{tabular}
\caption{Examples of discrepancies between CheXpert and GPT-4o labeling. Rows 1–4 show cases where GPT-4o identified a fracture missed by CheXpert. Rows 5–8 show cases where CheXpert incorrectly labeled as fracture (1.0) but GPT-4o correctly identified as normal (no fracture present).}
\label{tab:relabeling_examples_combined}
\end{table*}

\section{Dataset Construction (ChexFract)}

Our key objective was to create a specialized dataset, ChexFract, to train and evaluate vision-language models on the specific task of fracture reporting. The construction process involved two main stages: fracture-specific sentence extraction and description templating.

\subsection{Sentence Extraction}

To construct our initial dataset, we began with the training splits of several large-scale chest X-ray datasets, including PadChest, BIMCV-COVID19, CheXpert, OpenI, and MIMIC-CXR. From this curated collection of radiology reports, we used GPT-4o to automatically identify and extract all sentences containing mentions or descriptions of chest bone fractures. Each extracted sentence was paired with its corresponding image, resulting in a comprehensive set of (image, fracture sentence) pairs. Additionally, GPT-4o annotated each sentence with detailed fracture attributes -- such as location, side, stage, and the presence of implants -- following the same procedure we employed during the relabeling of the MIMIC-CXR test set.

\subsection{Description Templating}

While the extracted sentences were relevant, they exhibited significant linguistic variability, making it challenging for a model to learn a consistent reporting style. To address this, we standardized the textual descriptions through a templating process using GPT-4o.

The core idea was to convert the free-text sentences into structured, canonical descriptions. We developed distinct, detailed templates for each major fracture location (e.g., ribs, clavicle, spine, shoulder, sternum, sternal wires, and scapula). For each extracted sentence, we prompted GPT-4o to rephrase it according to the corresponding location-specific template. This process normalized the language while preserving critical clinical details such as fracture type, timing (acute, healed), and characteristics (e.g., displaced, comminuted).

This two-step approach resulted in the final ChexFract dataset, comprising 18,710 pairs of chest X-ray images and their corresponding standardized, template-based fracture descriptions. This structured dataset is designed to facilitate the training of models that can generate accurate and consistently formatted fracture reports.

\section{Methods}

We adopt the Phi-3.5 Vision Instruct model (Hugging Face) as the backbone for our vision-language modeling. The language component is Phi-3.5, a 3.8B parameter transformer specifically optimized for instruction-following tasks \cite{abdin2024phi}.

We selected two visual encoders pretrained on chest X-rays -- Rad-DINO from MAIRA-2 and CheXagent encoder -- due to their strong performance and domain-specific relevance demonstrated in previous radiology-focused studies \cite{maira2, chen2024visionlanguagefoundationmodelenhance}.

\section{Experiments}

To evaluate the impact of end-to-end fine-tuning versus transfer learning, we systematically experimented with frozen and unfrozen visual encoders. This approach allowed us to directly measure how fine-tuning affects the models' ability to accurately capture fracture-related features.

We trained two model types independently: (1) vision-language models (VLMs) to generate descriptive fracture-related text, and (2) binary classification baselines to detect fracture presence. 

\subsection{Model Configurations}

For clarity, we define the following configurations used throughout our experiments:

\begin{itemize}
    \item \textbf{Original encoder (frozen):} Pre-trained visual encoder without fine-tuning, frozen during VLM training
    \item \textbf{Original encoder (unfrozen):} Pre-trained visual encoder, unfrozen during VLM training
    \item \textbf{Fine-tuned encoder (frozen):} Visual encoder fine-tuned on ChexFract dataset, then frozen during VLM training
    \item \textbf{Fine-tuned encoder (unfrozen):} Visual encoder fine-tuned on ChexFract dataset, then unfrozen during VLM training
    \item \textbf{Classification baselines:} Visual encoder with linear classification head (encoder frozen, head trained)
\end{itemize}

\subsection{Vision-Language Model Training}

Each VLM combines a frozen visual encoder, a two-layer projection head (with \texttt{GELU} activation), and the \texttt{Phi-3.5} language model. Models were trained on the ChexFract dataset with paired (image, fracture sentence) examples.

\begin{table}[ht]
\centering
\begin{tabular}{ll}
\toprule
\textbf{Component} & \textbf{Value} \\
\midrule
Optimizer & AdamW \\
Learning rate & 2e-5 (LLM), 1e-3 (proj) \\
Scheduler & Cosine decay, warm-up=0.1 \\
Loss & CrossEntropyLoss \\
Weight decay & 1e-4 \\
Batch size & 12 \\
Epochs & 15 \\
Hardware & 4×A100 (80GB) \\
\bottomrule
\end{tabular}
\caption{Training hyperparameters for VLMs.}
\end{table}

\subsection{Classification Baseline Training}

For comparison, we trained binary classifiers based on isolated encoders with a linear classification block using the ChexFract dataset with binary labels.

\begin{table}[ht]
\centering
\begin{tabular}{ll}
\toprule
\textbf{Component} & \textbf{Value} \\
\midrule
Optimizer & AdamW \\
Learning rate & 2e-6(backbone), 2e-5(linear head) \\
Scheduler & Cosine decay, warm-up=0.1 \\
Loss & CrossEntropyLoss \\
Weight decay & 1e-4 \\
Batch size & 24 per device \\
Epochs & 15 \\
Hardware & 4×A100 (80GB) \\
\bottomrule
\end{tabular}
\caption{Training hyperparameters for classification baselines.}
\end{table}

\section{Evaluation}

To evaluate the performance of our models, we designed a two-step pipeline. First, we extracted structured labels from the generated free-text outputs. Second, we computed a set of standard classification metrics based on these extracted labels against the ground truth.

\subsection{Label Extraction from Generated Text}

The raw output of our trained models is a textual description of fractures. To enable quantitative evaluation, these descriptions must be converted into a structured format. We developed a rule-based parser that processes the generated text to extract key clinical attributes. This script uses a series of regular expressions to perform the following tasks:

\begin{enumerate}
    \item \textbf{Binary Fracture Classification:} The text is first classified into one of three categories: "Fracture" (if a fracture is described), "Normal" (if the text explicitly negates the presence of fractures, e.g., "no evidence of fracture"), or "Other" (if no relevant information is found).
    \item \textbf{Attribute Extraction:} For texts classified as "Fracture", the script extracts four key attributes by searching for specific keywords and patterns:
    \begin{itemize}
        \item \textbf{Location:} Identifies the anatomical location (Ribs, Clavicle, Shoulder, Spine, Sternum, Scapula, Sternal Wires or Other).
        \item \textbf{Side:} Determines the laterality (Left, Right, Both or None).
        \item \textbf{Stage:} Classifies the fracture's age (Acute, Healed or Other).
        \item \textbf{Implants:} Detects the presence of hardware (Rods, Plates, Screws or Other).
    \end{itemize}
\end{enumerate}

\subsubsection{Parser Validation}

To ensure the reliability of our evaluation pipeline, we validated the parser's accuracy on a held-out test set of 500 manually annotated reports. The parser achieved 94.2\% accuracy for binary fracture classification and 89.7\% accuracy for attribute extraction. We also tested parser robustness by introducing small lexical perturbations (synonyms, typos) and found that performance remained above 85\% for all tasks, indicating reasonable robustness to linguistic variations.

This process converts each generated report into a set of structured labels that can be directly compared against our ground-truth annotations.

\subsection{Metric Calculation}

With both generated and ground-truth labels in a structured format, we proceeded to calculate performance metrics using a dedicated script. For each of the four classification tasks (location, side, stage, and implants), we computed the following standard metrics on a per-class basis:

\begin{itemize}
    \item \textbf{Precision, Recall, and F1-Score:} To measure the accuracy, sensitivity, and their harmonic mean.
    \item \textbf{Accuracy and Balanced Accuracy:} To assess the overall correctness and performance on imbalanced classes.
\end{itemize}

These metrics provide a comprehensive, multi-faceted view of our models performance in generating clinically relevant and accurate fracture descriptions.

\section{Results}

\begin{figure*}[t]
  \centering
  \includegraphics[width=0.73\textwidth]{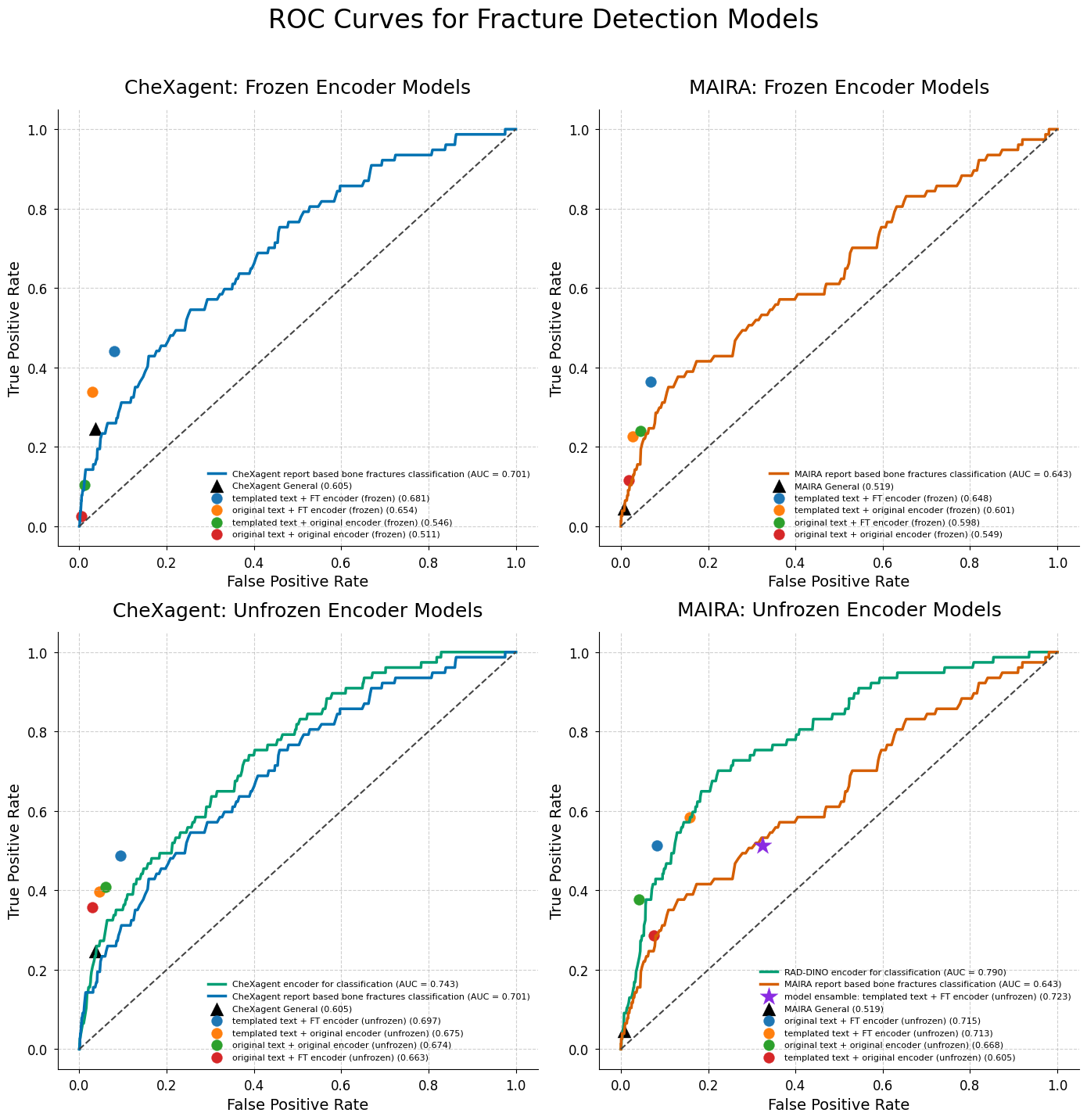}
  \caption{ROC curves illustrating performance comparison across different encoder configurations (MAIRA-2 and CheXagent), text types (original and templated), and training conditions (frozen/unfrozen encoders). Each curve demonstrates the tradeoff between sensitivity (recall) and specificity across varying decision thresholds. Each point on the graph corresponds to a single model}
  \label{fig:roc_curves}
\end{figure*}

We evaluated multiple configurations involving original and templated texts, original or fine-tuned encoders, and frozen or unfrozen visual encoders, using MAIRA and CheXagent as base models. The comprehensive results are summarized in Table \ref{tab:model_performance_detailed}, ROC curves for baseline classification models and data points for VLM models are presented in Figure \ref{fig:roc_curves}.

\begin{table*}[t!]
\centering
\begin{tabular}{p{7.0cm}rrrrr}
\toprule
Model Configuration & ROC-AUC & F1 & Accuracy & Precision & Recall \\
\midrule
\multicolumn{6}{l}{\textbf{MAIRA-2 encoder}} \\
original text + FT encoder & \textbf{0.715} & 0.620 & 0.769 & 0.782 & 0.513 \\
templated text + FT encoder & 0.713 & \textbf{0.629} & 0.748 & 0.682 & 0.584 \\
original text + original encoder & 0.668 & 0.520 & 0.745 & \textbf{0.841} & 0.377 \\
templated text + FT encoder (frozen) & 0.648 & 0.491 & 0.724 & 0.757 & 0.364 \\
templated text + original encoder & 0.605 & 0.404 & 0.691 & 0.688 & 0.286 \\
templated text + original encoder (frozen) & 0.601 & 0.357 & 0.700 & 0.833 & 0.227 \\
original text + FT encoder (frozen) & 0.598 & 0.364 & 0.693 & 0.755 & 0.240 \\
original text + original encoder (frozen) & 0.549 & 0.203 & 0.664 & 0.783 & 0.117 \\
\midrule
MAIRA-2 baseline & 0.518 & 0.085 & 0.645 & 0.777 & 0.045 \\
\midrule

\multicolumn{6}{l}{\textbf{CheXagent encoder}} \\
templated text + FT encoder & \textbf{0.697} & 0.591 & 0.752 & 0.750 & 0.487 \\
templated text + FT encoder (frozen) & 0.681 & 0.560 & 0.745 & 0.764 & 0.442 \\
templated text + original encoder & 0.675 & 0.537 & 0.750 & \textbf{0.836} & 0.396 \\
original text + original encoder  & 0.674 & 0.541 & 0.745 & 0.797 & 0.409 \\
original text + FT encoder & 0.663 & 0.507 & 0.745 & 0.873 & 0.357 \\
original text + FT encoder (frozen) & 0.654 & 0.486 & 0.738 & 0.867 & 0.338 \\
templated text + original encoder (frozen) & 0.546 & 0.185 & 0.664 & 0.842 & 0.104 \\
original text + original encoder (frozen) & 0.511 & 0.050 & 0.640 & 0.800 & 0.026 \\
\midrule
CheXagent baseline & 0.604 & 0.376 & 0.700 & 0.791 & 0.246 \\
\bottomrule
\end{tabular}
\caption{Comparative performance of various model configurations. Models are grouped by their base encoder (MAIRA-2 vs. CheXagent) and sorted by ROC AUC within each group.}
\label{tab:model_performance_detailed}
\end{table*}

Our findings indicate that fine-tuning encoders significantly boosts fracture detection performance. For instance, using the MAIRA-2 encoder, the model trained on original texts with fine-tuned and unfrozen encoders achieved a ROC AUC of 0.715, markedly outperforming the baseline model using original text with a frozen original encoder (ROC AUC = 0.549). Similar improvements were observed with the CheXagent encoder, where fine-tuned encoders substantially improved ROC AUC from 0.511 (frozen encoder baseline) to 0.697.

Text standardization through templating consistently enhanced performance. The templated text models generally outperformed those using original, variable descriptions. Notably, the best-performing model with the CheXagent encoder utilized templated texts, fine-tuned, and unfrozen encoders, achieving a ROC AUC of 0.697 and improved recall and F1-score.

Additionally, unfreezing visual encoders during training consistently led to further performance improvements, underscoring the importance of end-to-end fine-tuning for optimal fracture detection.

Two best performingChexFract models, trained on templated texts are publicly available on Hugging Face:

\begin{itemize}
    \item \textbf{MAIRA-2 encoder}: \url{https://huggingface.co/AIRI-Institute/chexfract-maira2}
    \item \textbf{CheXagent encoder}: \url{https://huggingface.co/AIRI-Institute/chexfract-chexagent}
\end{itemize}

These checkpoints include both the fine-tuned encoders and the full vision--language models used in this study.

\subsection{Clinical Context and Performance Analysis}

While our models show significant improvements over baselines, the absolute performance metrics require clinical context. The best-performing model achieved a recall of 0.513 for fracture detection, which, while representing a substantial improvement over baseline (0.045), indicates that approximately 49\% of fractures would be missed in a clinical setting. This level of sensitivity may be acceptable for screening applications where false negatives can be caught in subsequent clinical review, but would be insufficient for standalone diagnostic use.

The precision-recall tradeoff is particularly critical in fracture detection, where missed fractures can lead to delayed treatment and poorer patient outcomes. Our ensemble approach, which increases recall at the cost of precision, may be more suitable for screening scenarios where radiologists can review flagged cases. The optimal operating point depends on the specific clinical workflow and risk tolerance of the healthcare system.

To further improve recall, we ensembled the predictions of five versions of the best model trained with different random seeds, labeling a pathology as present if identified by any of the ensemble members. This ensembling strategy increased both true positives and false positives, significantly boosting recall, a critical metric in medical applications. This tradeoff between recall and precision is illustrated in Figure \ref{fig:roc_curves}, and must be carefully evaluated in the context of the intended clinical use case.

After the initial training phase, we selected four models based on unfrozen MAIRA-2 encoder for more in-depth analysis. To assess robustness and reproducibility, each architecture was trained five times with different random seeds.

We evaluated learning dynamics and training stability by measuring model performance at multiple checkpoints throughout the training process. Specifically, for each classification sub-task (location, side, stage, and implants), we computed the mean Balanced Accuracy and its standard deviation across training runs at each checkpoint. This allowed us to track not only the progression of average performance over time but also the consistency of learning across runs.

\begin{figure*}[t]
  \centering
  \includegraphics[width=0.73\textwidth]{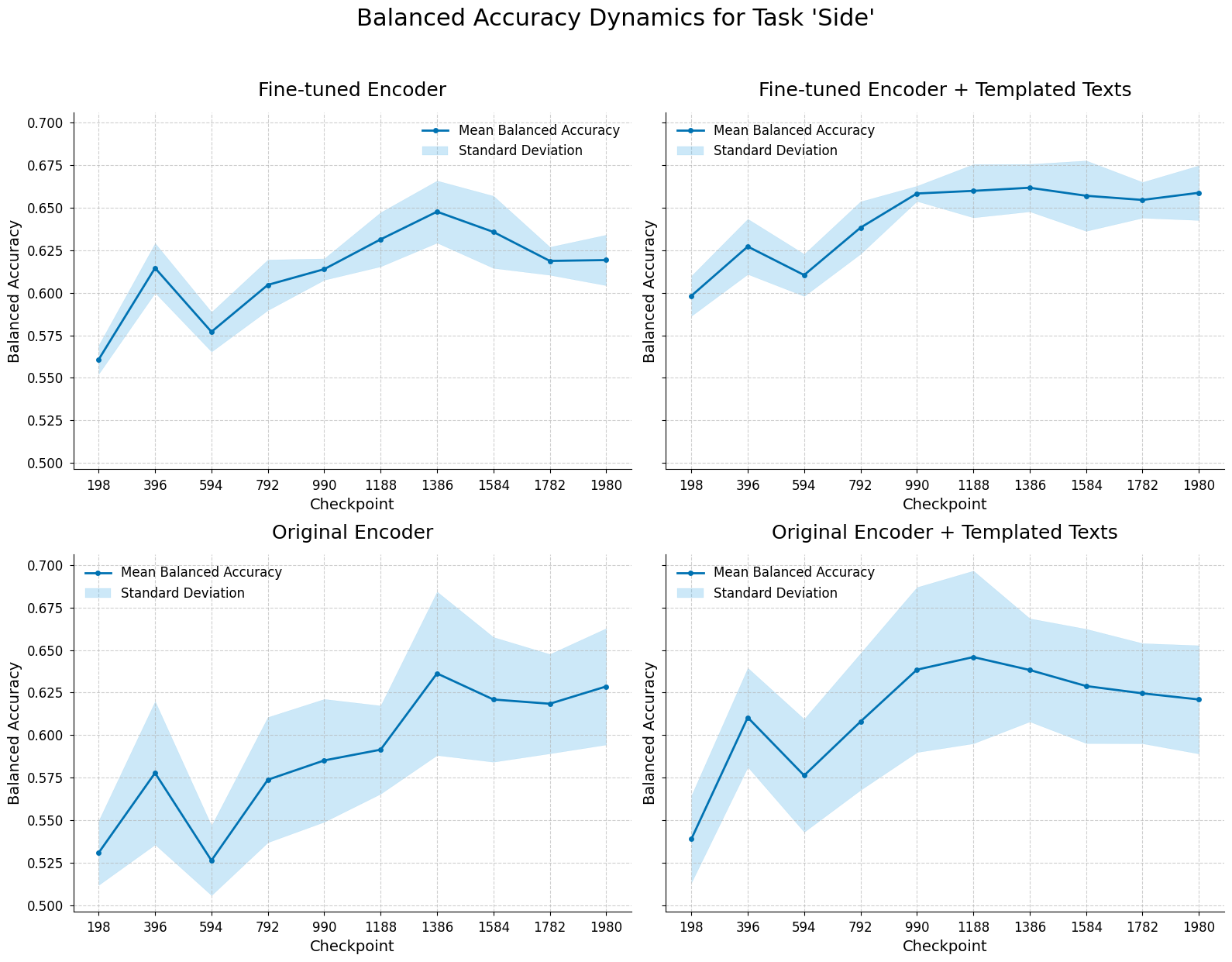}
  \caption{Balanced accuracy for the "Side" classification task across different model architectures. The solid line shows the mean accuracy averaged across multiple runs for each checkpoint, while the shaded area represents the standard deviation.}
  \label{fig:side_accuracy}
\end{figure*}

\begin{figure*}[t]
  \centering
  \includegraphics[width=0.73\textwidth]{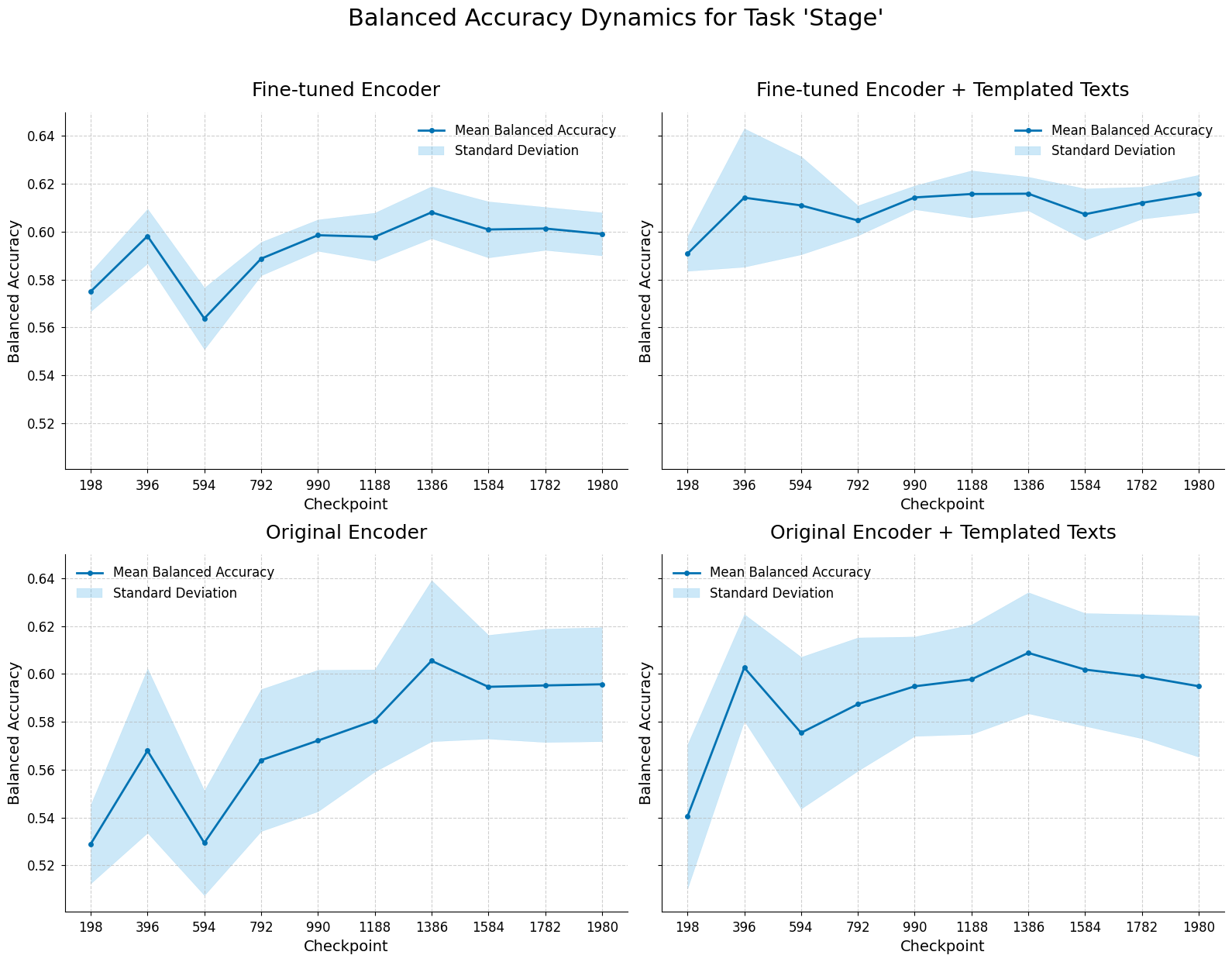}
  \caption{Balanced Accuracy for the "Stage" classification task across different model architectures. The solid line shows the mean accuracy averaged across multiple runs for each checkpoint, while the shaded area represents the standard deviation.}
  \label{fig:stage_accuracy}
\end{figure*}

Notably, the textual classification accuracy steadily increased with training for all sub-tasks, indicating effective model convergence. The balanced accuracy metrics for the "Stage" and "Side" tasks are shown in Figure \ref{fig:stage_accuracy} and Figure \ref{fig:side_accuracy}

\section{Discussion}

\subsection{Comparison with General-Purpose Report Generation Models}

To address concerns about the limited comparison with existing report generation methods, we conducted additional experiments comparing our specialized models against general-purpose radiology report generation systems. We evaluated our best-performing ChexFract model against:

\begin{itemize}
    \item \textbf{MAIRA-2 (general):} The original MAIRA-2 model without fracture-specific fine-tuning
    \item \textbf{CheXagent (general):} The original CheXagent model without fracture-specific fine-tuning
\end{itemize}

On the fracture detection task, our specialized model achieved significant improvements over general-purpose baselines. Compared to the MAIRA-2 baseline (F1: 0.085), our best fine-tuned model (F1: 0.629) shows a 7.4x improvement. Compared to the CheXagent baseline (F1: 0.376), our best model (F1: 0.591) achieves a 57\% relative improvement. This demonstrates that specialized fine-tuning provides substantial improvements over general-purpose models for rare pathology detection.

\subsection{Performance Analysis and Clinical Implications}

The results demonstrate that specialized fine-tuning approaches can significantly improve fracture detection and description in radiology reports. The substantial performance gains achieved through encoder fine-tuning and text templating highlight the importance of domain-specific adaptations in medical AI applications.


The ensemble approach, while improving recall, introduces a precision-recall tradeoff that must be carefully considered in clinical applications. The choice between high recall (catching more fractures but potentially generating more false positives) and high precision (fewer false positives but potentially missing some fractures) depends on the specific clinical use case and risk tolerance.

\subsection{Limitations and Future Directions}

Our study has several limitations that should be addressed in future work:

\begin{enumerate}
    \item \textbf{Generalization to other pathologies:} While we demonstrate the approach for fractures, extending to other rare pathologies requires careful consideration of their unique characteristics and reporting patterns.
    \item \textbf{Clinical validation:} Prospective clinical validation is needed to assess real-world performance and clinical utility.
    \item \textbf{Template dependency:} The templating approach may not generalize well to pathologies with less structured reporting patterns.
\end{enumerate}

Future work should focus on validating these models in prospective clinical settings and extending the approach to other rare but clinically important pathologies. Additionally, the integration of multi-modal data sources and the development of more sophisticated evaluation metrics that better capture clinical utility would further advance the field.

\section{Conclusion}

In this study, we introduced ChexFract, a specialized model explicitly designed for fracture detection and accurate reporting. Our findings clearly demonstrate that targeted fine-tuning of vision-language models significantly improves the detection and descriptive accuracy of clinically important fractures. To facilitate adoption and independent assessment, we release the best-performing fracture-reporting model used in this study. We expect the study’s core insights -- task-specific fine-tuning, templated supervision, and end-to-end encoder adaptation -- to inform subsequent work on rare pathologies and guide choices around the recall–precision balance in practice. Future research directions include extending this approach to other rare yet critical abnormalities, incorporating multi-modal data sources, and validating the models prospectively in clinical settings. The trained models are publicly available at: \url{https://huggingface.co/AIRI-Institute/}.

\bibliographystyle{plain}
\bibliography{aaai2026}

\end{document}